\documentclass[journal]{IEEEtran}
\ifCLASSINFOpdf
\else
\fi

\usepackage{color}
\usepackage{url}
\usepackage{graphicx}
\usepackage{multirow}
\usepackage{array}
\usepackage[english]{babel}
\usepackage{caption2}

\usepackage{algorithm}
\usepackage{algpseudocode}
\usepackage{amsmath}

\usepackage{color} 
\begin{document}

\title{A Real-time Action Representation\\ with Temporal Encoding and Deep Compression}
\author{Kun~Liu,~\IEEEmembership{Student Member,~IEEE,}
        Wu~Liu,~\IEEEmembership{Member,~IEEE,}
        Huadong~Ma,~\IEEEmembership{Senior Member,~IEEE,}

        Mingkui~Tan,~\IEEEmembership{Member,~IEEE,}
        and~Chuang~Gan,~\IEEEmembership{}
\IEEEcompsocitemizethanks{\IEEEcompsocthanksitem

K.Liu, W.Liu, and H. Ma are with the Beijing Key Laboratory of Intelligent Telecommunication Software and Multimedia, Beijing University of Posts and
Telecommunications, Beijing 100876, China (e-mail: liu\_kun@bupt.edu.cn;
liuwu@bupt.edu.cn; mhd@bupt.edu.cn).

M.Tan is with South China University of Technology, Guangzhou, China (e-mail: mingkuitan@scut.edu.cn)

C.Gan is with MIT-Watson AI Lab, Cambridge, Massachusetts, United States of America (e-mail: ganchuang1990@gmail.com).
\IEEEcompsocthanksitem Color versions of one or more of the figures in this paper are available online at http://ieeexplore.ieee.org.}
\thanks{}}

\markboth{IEEE TRANSACTIONS ON CIRCUITS AND SYSTEMS FOR VIDEO TECHNOLOGY}
{Shell \MakeLowercase{\textit{et al.}}: Bare Demo of IEEEtran.cls for IEEE Journals}

\maketitle
\begin{abstract}
Deep neural networks have achieved remarkable success for video-based action recognition.
However, most of existing approaches cannot be deployed in practice due to the high computational cost.
To address this challenge, we propose a new real-time convolutional architecture, called Temporal Convolutional 3D Network (T-C3D), for action representation.
T-C3D learns video action representations in a hierarchical multi-granularity manner while obtaining a high process speed.
Specifically, we propose a residual 3D Convolutional Neural Network (CNN) to capture complementary information on the appearance of a single frame and the motion between consecutive frames.
Based on this CNN, we develop a new temporal encoding method to explore the temporal dynamics of the whole video.
Furthermore, we integrate deep compression techniques with T-C3D to further accelerate the deployment of models via reducing the size of the model.
By these means, heavy calculations can be avoided when doing the inference, which enables the method to deal with videos beyond real-time speed while keeping promising performance.
Our method achieves clear improvements on UCF101 action recognition benchmark against state-of-the-art real-time methods by 5.4\% in terms of accuracy and 2 times faster in terms of inference speed with a less than 5MB storage model.
We validate our approach by studying its action representation performance on four different benchmarks over three different tasks.
Extensive experiments demonstrate comparable recognition performance to the state-of-the-art methods.
The source code and the pre-trained models are publicly available at \url{https://github.com/tc3d.}
\end{abstract}
\begin{IEEEkeywords}
Real-time, action recognition, temporal encoding, deep compression
\end{IEEEkeywords}
\IEEEpeerreviewmaketitle

\section{Introduction}

\IEEEPARstart{T}{he} representation of human actions, which tries to capture the powerful features of action,
is an important research topic in the video understanding community.
Video-based action recognition aims to make computers recognize human actions automatically in real-world videos.
Many researchers in video understanding and computer vision fields have concentrated on action representation due to its widespread application,
such as video classification \cite{peng2019two,zhang2014cross} and event detection \cite{xian2016evaluation,liu2018multi}.
The task of human action recognition in videos, however, is still very challenging due to the three following reasons.

\begin{figure}[t]
\begin{minipage}[b]{0.96\linewidth}
  \centering
  \centerline{\includegraphics[width=9.0cm,height=7.2cm]{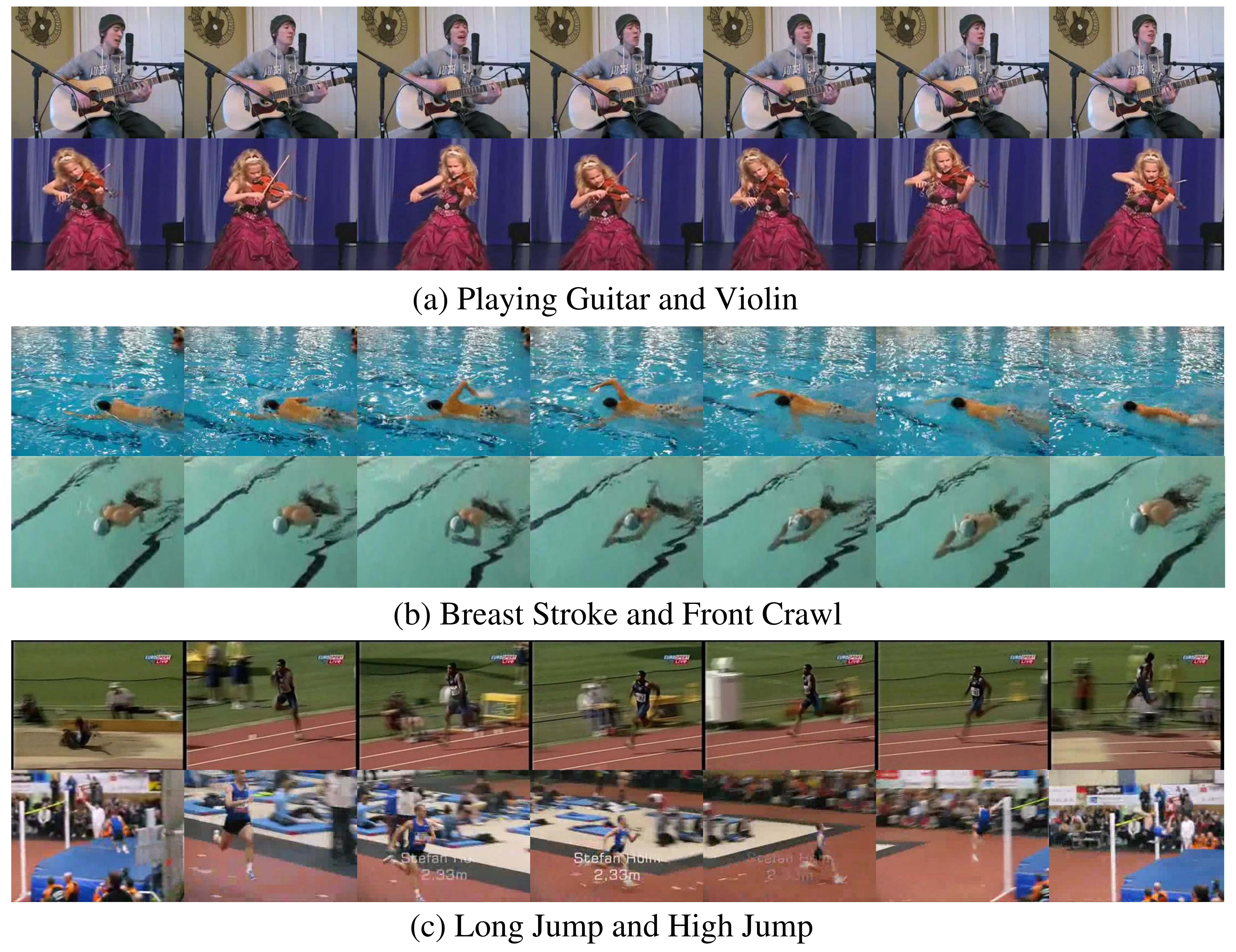}}
\end{minipage}
\centering \caption{Example videos from six categories of UCF101. Some actions can be reliably distinguished through a single frame or the motion computed from consecutive frames. However, certain similar actions such as ``High Jump'' and ``Long Jump'' require long-term features of the video, since the short-term information on a clip is almost the same.}
\label{fig:01}
\end{figure}
First, the video is naturally an information-intensive media with hierarchical multi-granularity structure, e.g., static information in one frame, short-term slow motions in a clip, and long-term temporal evolution of appearances.
As shown in Figure \ref{fig:01}(a), some actions, such as ``playing guitar'' and ``playing archery'', can be accurately recognized through extracting features of only one frame.
Meanwhile, a number of human events require more abundant information like the motion between several consecutive frames to identify the action.
For example, Figure \ref{fig:01}(b) demonstrates that ``front crawl'' and ``breaststroke'' can be differentiated via one clip rather than one frame.
As for some highly similar human actions, as shown in Figure \ref{fig:01}(c), several clips cannot precisely distinguish both actions due to the almost identical visual representation, especially during the run-up phase in ``high jump'' and ``long jump''.
Therefore, we suppose that efficiently learning a multi-level temporal structure is an indispensable part of recognizing actions in videos.
The above argument not only is consistent with human intuition,
but also has been validated in~\cite{Huang2018What} through analyzing motion information in videos.

To deal with the above challenges, we introduce an effective and efficient architecture, named T-C3D, by integrating the deep residual 3D network with the temporal encoding. The 3D convolutional network is able to capture both the static image information and the short-term motion features in successive frames.
The long-term information is characterized by the temporal encoding algorithm.
As a result, our T-C3D can represent the hierarchical multi-granularity structure of the actions,
achieving competitive performance compared with the state-of-the-art methods.

Second, the calculation of most methods for action representation is too high to deploy in practice.
These methods can be fallen into two categories:
hand-crafted descriptor-based methods ~\cite{klaser2008spatio,wang2013action,wu2013action}
and action representations with deep learning~\cite{karpathy2014large,simonyan2014two,kar2016adascan,tran2015learning,tran2017convnet,varol2017long,qiu2017learning,girdhar2017actionvlad}.
The former predominantly consists of feature extraction, feature encoding,
and classification, which mainly focuses on developing discriminative and powerful spatial-temporal video descriptors.
Different from the methods with hand-crafted descriptors, most deep learning methods make use of Convolutional Neural Networks (CNNs) and Recurrent Neural Networks (RNNs) to exploit individual image-level appearance and model the temporal characteristic in videos.
However, the majority of the impressive methods, such as improved Dense Trajectory (iDT)~\cite{wang2013action}, two-stream~\cite{simonyan2014two} and their improved version ~\cite{kar2016adascan,girdhar2017actionvlad,wang2016temporal}, cannot meet the real-time requirement due to their high computation cost, e.g., the calculation of optical flow.
Unfortunately, the application of action recognition in real life is severely impeded by this obstacle.

To circumvent this difficulty, we propose to accelerate the deployment of our T-C3D from three aspects.
First, we adopt the three-dimensional convolutional neural network to obtain the motion between continuous frames, which significantly avoids the heavy calculation compared with the two-stream based methods as they extract the optical flow.
We also adopt some efficient techniques to prevent over-fitting when training the T-C3D.
Moreover, during the training process, we design the temporal encoding algorithm which aggregates the feature of several clips sampled from the same video,
which leads to a promising accuracy when fusing only a small fraction of testing clips rather than predicting all of the testing instances.
This also dramatically decreases the inference time.
More importantly, we speed up the T-C3D from weight distribution.
More specifically, we first progressively prune all layers of T-C3D to set most values of the weight to be zero.
Then we store the pruned sparse results using Compressed Sparse Row (CSR) format.
In the last step, we reduce computational complexity via a certain kernel designed for optimizing the sparse matrix-vector multiplication on popular GPUs.

Third, the existence of large video benchmarks for action recognition and worldwide challenges, like UCF101 \cite{soomro2012ucf101}, Sports-1M \cite{karpathy2014large}, and Kinetics \cite{kay2017kinetics} has boosted researches in this area.
Consequently, powerful models have mainly been appraised according to classification accuracy.
Meanwhile, most work in the video understanding community pays little attention to
computational efficiency and model complexity for applying action recognition into the real-world.
For example, the C3D model \cite{tran2015learning} saved in a format of caffemodel is over 300MB,
which consumes expensive memory storage and results in the difficulty of application.
While video-based action recognition has made great achievements in terms of classification accuracy,
successful deployment of action recognition in real-life, which is beneficial to social development, is still missing.
The major obstacle for the deployment is the requirement of small model sizes to accommodate limited memory on the smart devices.
Many applications on mobile devices are extremely sensitive to the size of the files.
For instance, App Store has the restriction ¡°apps above 100 MB will not download until you connect to Wi-Fi¡± \cite{han2015deep}.
Therefore, apps with fewer model sizes are more likely to be installed and updated.
This motivates us to concentrate on reducing the model size and inference time of T-C3D while preserving the accuracy.

To relieve this problem, we compress the model size of T-C3D via three following strategies: 1) prune all layers of the network, 2) quantize the weights of the model, and 3) encode the model weights with Huffman coding.
More specially, we sort the T-C3D weights and set the weights with less value to be zero, obtaining a sparse model whose most values are zero.
To utilize fewer bits to represent more weights, we cluster each layer of T-C3D via K-means.
In the last step, we adopt the Huffman coding, a particular type of optimal prefix code designed for lossless data compression, to further compress the model size without sacrificing the performance.
On the UCF101 dataset, these above measures significantly reduce the storage requirement of T-C3D from 128MB to 4.2MB, almost without loss of accuracy.

To validate the effectiveness of our proposed method, extensive experiments are conducted on four datasets over three tasks, including action recognition, action similarity labeling, and anomaly detection in surveillance videos.
In particular, on the action recognition task, we appraise the accuracy, speed and model size of proposed T-C3D.
Experiments demonstrate that T-C3D outperforms the-state-of-the-art methods by 5.4\% in terms of accuracy and 2 times faster in terms of speed with a less than 5MB storage model.
On both the action similarity labeling and anomaly detection, T-C3D also improves the existing methods by a substantial margin.
In particular, T-C3D achieves superior performance against the C3D from 50.0\% to 76.26\% on the anomaly detection benchmark in terms of Area Under Curve (AUC).

We extend our previous work \cite{tc3d2018liukun} from three aspects.
First, we integrate a serious of deep compression techniques with temporal encoding algorithm to
further accelerate the employment of this framework.
In terms of performance, deep compression techniques reduce the model size from
230MB to less than 5MB without accuracy degradation.
Second, we validate our method on another two benchmarks over two different tasks.
More specially, we adopt our method to detect anomaly in surveillance videos to
verify that our method can be deployed in practice.
Third, we provide further insight into our learned models via some subjective results, such as visualization.
\newline
\newline
In summary, the main contributions of this work are summarized as follows:
 \begin{itemize}
    \item We propose a 3D-CNN based framework equipped with the temporal encoding method to learn action representation at a multitude of granularity. 
    \item We introduce several compression techniques to reduce the original model size of T-C3D to less than 5MB, which significantly promotes the application of our T-C3D into the real-world.
    \item Our framework can process videos at real-time while achieving competitive accuracy since we avoid computationally expensive steps to extract video features.
\end{itemize}

\section{Related Work}
\label{relat}
Action representation has been extensively studied in the past few years.
We classify major prior work related to ours into two classes: 1) action recognition with hand-crafted features, and 2) action recognition with deep learning.

\subsection{Action Recognition with Hand-crafted Features.} To capture the intrinsic temporal motion in videos, extensive work makes efforts to manually develop powerful features for action recognition.
This kind of approach always follows these procedures: feature extraction, feature encoding, and classification.
Researchers often focus on designing powerful spatial-temporal video descriptors which are developed from the static single image area and extended to depict the three dimension of videos (temporal dimension), including 3D Histogram of Gradient (HOG3D) \cite{klaser2008spatio} and 3D Scale-Invariant Feature Transform (SIFT-3D) \cite{scovanner20073}.
Moreover, some work concentrates on devising local spatio-temporal features to model the temporal dimension.
Besides, Wang et al. \cite{wang2013action} propose a powerful hand-crafted feature named Improved Dense Trajectories (IDT),
which extracts several descriptors (HOG, HOF, and MBH) and tracks them in a dense optical flow field.
Most of the features designed by human beings are not discriminative enough to obtain satisfactory performance.
While few hand-crafted features exhibit the promising experimental results, e.g., IDT,
these methods cannot meet the requirement of real-time due to the expensive computation cost caused by the calculation of optical flow.

\subsection{Action Recognition with Deep Neural Network.} Since deep neural networks bring remarkable achievements over many visual tasks, such as video summarization \cite{wang2012event,wang2012movie2comics} and video retrieval \cite{liu2013listen,liu2017deep}.
This sort of approaches make use of neural networks to learn action representation automatically in an end-to-end manner,
which eliminates the inaccuracies of manual design.
According to different network architecture, deep neural networks for action recognition can fall into two categories: one-stream framework based methods and two-stream architecture based approaches.

\textbf{One-stream framework.} As to the one-stream framework \cite{tran2015learning,tran2017convnet,gan2015devnet,huang2018toward,gan2016recognizing,gan2018geometry,liu2018learning} , networks are always designed to capture both the spatial feature and the temporal information simultaneously.
Some work \cite{tran2015learning,tran2017convnet,varol2017long} utilizes 3D convolution filters with the input of a short snippet to capture the temporal dimension motion in videos.
Recently, Qiu et al.\cite{qiu2017learning} factorize the standard 3D convolutional filters into 1D temporal filters and 2D spatial kernel and achieve competitive performance.
However, this kind of approach has not yet dramatically exceeded the state of the art hand-crafted methods for video-based action recognition.

We argue there are two reasons for this phenomenon.
First, this is partly owing to the lack of the ability to grasp the characteristics of long-term features.
Obviously, feeding 3D-CNN with more consecutive frames can capture longer temporal features.
However, this strategy is largely limited by GPU memory and obtains slight improvements \cite{varol2017long}.
The other reason might be the failure to unleash the full potential of large-scale video benchmarks comparable in size and diversity with ImageNet,
which can avoid over-fitting.
The emergence of large-scale and well-labeled benchmarks for action recognition, like Sports-1M \cite{karpathy2014large} and Kinetics \cite{kay2017kinetics},
bring the opportunities to advance researches in this domain.
To tackle these above difficulties, we not only integrate the standard 3D-CNN with temporal encoding algorithm to model the long-term temporal character
but also perform initialization of T-C3D parameters on large-scale video datasets to fully inspire the potential of 3D-CNN.

\begin{figure*}[t]
\begin{minipage}[b]{0.96\linewidth}
  \centering
  \centerline{\includegraphics[width=14.6cm,height=6.5cm]{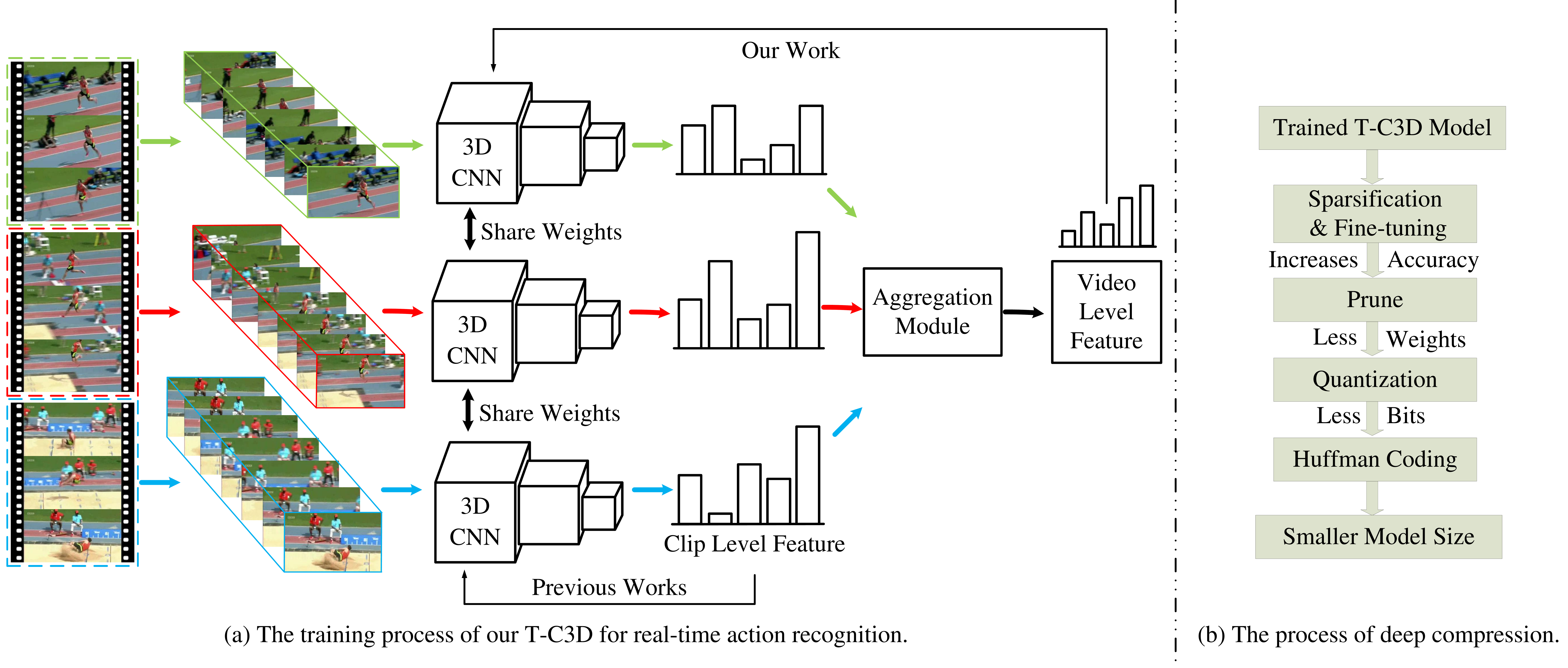}}
\end{minipage}
\centering \caption{The framework of the proposed architecture for real-time video action representation.}
\label{fig:02}
\end{figure*}

\textbf{Two-stream framework.} The two-stream approach for action recognition is first proposed in \cite{simonyan2014two}, in which spatial network acquires single RGB frame feature and temporal network captures the motion pattern between frames with the input of $10$ stacked optical flow images.
Then a great number of efforts \cite{girdhar2017actionvlad,wang2016temporal,wang2017first} have been made to enhance the two-stream network from different perspectives.

Since RNN has exhibited remarkable ability on sequence modeling, some researchers \cite{yue2015beyond,li2016action}incorporate the two-stream framework with RNN to capture the long-term information.
Recently, Kar et al. \cite{kar2016adascan} propose an adaptive temporal pooling method that selects informative frames and ignores the unnecessary and non-discriminative frames in the video.
The method described in \cite{girdhar2017actionvlad} offers a novel feature aggregation scheme for two-stream in an end-to-end trainable manner,
where both the feature extractor (Two-stream CNNs) and the inherent visual vocabularies (VLAD) are learned automatically by minimizing the loss value.

Despite outstanding performance, all of the two-stream based approaches are too computationally expensive to process videos in real-time because of the heavy calculation of optical flows.
Besides, during the testing stage, two separate models require to be deployed, which makes the application of action recognition in real life more difficult.

For real-time action recognition, Zhang et al. \cite{zhang2016real} propose a real-time algorithm that utilizes the enhanced motion vectors to replace the optical flow.
Although this work indeed speeds up the architecture of deep neural networks for video-based action representation,
it is cumbersome owing to the calculation of optical flow during the training phase and
the deployment of two models (RGB model and motion vector model) during the testing stage.
Compared with it, our approach fully prevents extracting the optical flow and only adopts video frames to train the network.
Meanwhile, our method can achieve a higher classification accuracy at a faster speed.
\section{The Proposed Method}
\label{Proposed}
\subsection{Overview of T-C3D Framework}
\label{Overv}
Figure \ref{fig:02}(a) shows the training process of the proposed T-C3D network.
First of all, we divide every given input video into $S$ parts according to the temporal dimension.
Then dozens of frames are chosen from each part to constitute a clip.
Next, $S$ clips sampled from the same video are fed into the 3D-CNNs simultaneously.
Then, the feature maps or class possibilities of $S$ clips are aggregated by a fusion function to generate the consensus,
which is a video-level prediction since it integrates multiple clips selected from the entire video.
In the last step, our T-C3D optimizes the neural network by calculating the loss value with the video-level score.

As a result, T-C3D can capture both short-term and long-term features in videos.
First, the 3D-CNN captures the short-term information between continuous frames.
Especially, the 3D-CNN expands the original 2D convolutional kernel at the temporal dimension,
which might be more applicable to capture the intrinsic three-dimensional feature in videos.
However, these 3D convolutions require dense sampling in time to work effectively.
Consequently, to fit in GPU memory the video inputs must be short.
Therefore, we design the temporal encoding model to extract long-term features.
Especially, the video-level predictions which integrate several clips' information are adopted to update
the parameters of networks.
In contrast, most of the previous work calculates the loss with clip-level or single frame prediction, ignoring the critical long-term information.
Furthermore, both 3D-CNN and temporal encoding model are trained in an end-to-end manner.

In brief, the framework is mainly made up by the following procedures: 1) producing the clips of each video,
2) feeding the clips to the 3D network simultaneously, 3) generating the video-level predicted possibilities by fusing the features of the clips,
and 4) updating the weights of 3D-CNN using the video-level prediction.
All the above steps are end-to-end trainable.
Then, we will present the proposed temporal encoding algorithm and every step in detail.
Finally, deep compression techniques for reducing the model size are introduced.

\subsection{Temporal Encoding Model}
\label{Tempo}
The video is a kind of intrinsically hierarchical structured medium since a video can be parsed into a spatial static image,
temporal parts are formed by short-term motion among continuous frames as well as long-term temporal evolution in videos.
Inspired by this observation and considering the real-time demand,
we introduce a novel real-time framework to recognize the actions in videos.
First of all, we replace computational expensively features (e.g., optical flow) with 3D convolutional filters to depict the short temporal motion,
which brings the framework a sharp acceleration owing to the only light-weight extraction of video frames.
Next, we introduce the temporal encoding algorithm to capture the long-term temporal features,
which can significantly increase the recognition performance.

To capture the long-term temporal features, we develop the temporal encoding algorithm.
Formally, given an input video $V$ during the training stage, we parse it into $S$ parts \{$P_1, P_2, P_3, ..., P_s$ \} at temporal dimension.
Next, dozens of frames are selected from $Pi$ to constitute the clip $Ci$.
Then, we feed the 3D-CNN all clips at the same time to yield $S$ feature maps or category possibilities.
Next, $S$ clips features are fused with specific aggregating functions to produce the video-level features.
Finally, video-level features derive the final category possibilities by forwarding a few convolutional or fully-connected layers.
Compared with previous methods, T-C3D updates its network weights via the video-level prediction rather than the single frame or one clip score.
The whole procedure can be formulated as Equation \ref{equ:01}:
\begin{equation}
Y_v=H(Q(F(C_1;W);F(C_2;W); ...; F(C_s;W))),
\label{equ:01}
\end{equation}
where $Y_v$ stands for the ultimate category possibility of the video $v$,
$F(C_s,W)$ is the function that represents the 3D-CNN with weights $W$ and produces feature map of clip $C_s$,
such as last convolutional layer, fully-connected layer, and final scores of all action classes.
The fusion function $Q$ aggregates the feature maps of multiple clips to generate a powerful representation for the long-term temporal character in videos.
Based on this discriminative representation, the score function $H$ generates the score values of every action class for the entire video.
In this work, by default, the output of $F$ is the last fully-connected layer.
$H$ is the standard softmax function to get the probability distribution.
We explore several choices of fusion function $Q$ in Section \ref{exper:4}.

The differentiability of the proposed temporal encoding approach makes the network easier to train with the widely applied back-propagation.
Through original categorical cross-entropy loss, the ultimate loss functions concerning the multiple clip's consensus
$G=Q(F(C_1;W);F(C_2;W); ...; F(C_s;W))$ can be formulated as
\begin{equation}
    L(y,G)= -\sum_{i=1}^{N}{y_i}(G_i-\log{\sum_{j=1}^{N}{\exp {G_j}}}),
\label{equ:02}
\end{equation}
where $N$ is the total of all action categories, $y_i$ is the ground-truth label regarding category $i$ and $S$ is a hyper-parameter.
The sensitivity study of $S$ is conducted in Section \ref{exper:4}.

In our method, the fusion function $Q$ plays a significant role since it not only aggregates several clips' feature into a video-level score
but also determines the differentiability of the entire framework.
The main difference between our temporal encoding method and previous methods is to
optimize the network via the video-level score rather than the features of clips.
It is aggregation functions that fuse the features of clips into the video-level score.
In this work, we extensively explore several differentiable fusion function alternatives, e.g., average pooling, maximum pooling, and attention pooling.
Differentiable fusion functions allow us to adopt the multiple clips to jointly optimize the network parameters $W$ through back-propagation algorithms.
Formally, during the back-propagation procedure, the gradients of T-C3D weights $W$ regarding the loss value $L$ can be formulated as
\begin{equation}
    \frac{\partial L(y,G)}{\partial W}= \frac{\partial L}{\partial W}\sum_{s=1}^{S}\frac{\partial Q}{\partial F(C_s)} \frac{\partial F(C_s)}{\partial W},
\label{equ:03}
\end{equation}
in which $S$ is number of clips.
In Equation \ref{equ:03}, the weights are updated via the multiple clips' consensus $G$ derived from all clip-level scores.
Optimized network parameters in this manner, T-C3D trains network weights from long-term temporal features instead of a single clip.
Then, we will describe the implementation of the model in detail.

\subsection{Video Components Generation}
\label{sbusec:VideoComponentsGeneration}
 Compared with the static images, videos are dynamic and consist of varying sequences.
 To explore the proper means to obtain the long-term temporal information, we first parse the video into a few parts at the temporal dimension.
 Then one clip is formed by selecting a certain number of frames from every part with two sampling strategies.
 The first strategy uniformly divides the video snippet generated in the previous step into a certain number of fragments.
 Then one frame is randomly selected from each fragment to form the clip.
 The second scheme randomly selects the specified number of consecutive frames from the snippet to constitute the clip.
 In essence, the first sampling scheme randomly chooses non-consecutive frames distributed uniformly throughout the entire video.
 The second approach evenly selects $S$ clips from the whole video and each clip is composed of the specified amount of consecutive frames.

 Formally, given a video consisted of $N$ frames, we first cut the video into $S$ pieces averagely at the temporal dimension,
 and therefore each part contains $\frac{N}{S}$ frames. For simplicity, we annotate $\frac{N}{S}$ as $M$.
Then, both sampling strategies form the a clip with $k$ frames according to each piece:
 \begin{equation}
 Clip =   \{ f_{1},f_{2},f_{i},...,f_{k} \},
 \label{equ:add01}
\end{equation}
 the frame number of $f_{i}$ is random(0,$\frac{M}{k}$)+$\frac{M}{k}$*$i$ for the first sampling strategy,
 while the frame number of $f_{i}$ of the second scheme is random(0,$M-k*o$)+$i*o$.
 $o$ means the offset between two sampled frames.
 The clip generated by the second strategy is composed of consecutive frames as the $o$  has nothing with $M$ and $i$.
 In our experiment, ${S}$ is 3, $k$ is 8, $o$ is 2.

For example, given a video with 300 frames, the first scheme may generate three
segments with the following frame number: \{\{1, 23, 26, 40,
53, 68, 81, 88\}, \{110, 120, 134, 143, 160, 164, 179, 185\},
\{204, 213, 231, 238, 249, 268, 282, 295\}\}, while the second
may produce three segments with the following frame number:
\{\{3, 5, 7, 9, 11, 13, 15, 17\}, \{125, 127, 129, 131, 133,
135, 137, 139\}, \{251, 253, 255, 257, 259, 261, 263, 265\}\}.
Obviously, the clips selected by the first strategy may contain
large and imbalanced intervals. In the next section, we compare
the recognition accuracy of both different sampling schemes.

\subsection{3D Convolutional Neural Network}
Given 2D-CNN's remarkable performance on a variety of vision tasks in the image domain,
3D-CNN is likely to achieve success in the video area since it can be regarded as the expansion of images at the temporal dimension.
Convolutional 3D Network (C3D) \cite{tran2015learning} is one of the first classical work that adopts the 3D convolutional filter to model both the spatial appearance and temporal information with the input of $16$ frames.
However, sixteen consecutive RGB frames do not involve long-term cues.
Thus, Long-term Temporal Convolutions (LTC) \cite{varol2017long} enhance the C3D through feeding network with more continuous RGB frames as long as the GPU memory allows, ranging from 20 to 100 frames.
The aforementioned work proves that 3D-CNN is a promising direction to recognize the actions in videos.
In this paper, we extend the above 3D-CNN work from the following perspectives.

First of all, inspired by the excellent image recognition performance obtained by the 2D-CNN with residual module \cite{he2016deep},
we design a deep 3D-CNN architecture with a powerful residual block.
More specially, according to the previous study on 3D-CNN architecture search \cite{tran2017convnet},
we employ the 3D-CNN with seventeen 3D convolutional layers and one fully-connected layer.
Experimental results show that the deeper 3D-CNN with residual block can learn more discriminative and robust spatiotemporal features from the given clips.

Secondly, initializing the weights of CNN on a large-scale dataset has been demonstrated greatly important for many visual topics\cite{liu2015multi}. As for 3D-CNN, previous work such as LTC has shown that 3D networks pre-trained on a large-scale but not well-labeled dataset achieve better recognition results than the models trained from scratch.
In our work, we first follow the scheme described in C3D and pre-train our network on Sports-1M.
While Sports-1M contains more than one million videos, part of its ground-truth labels are wrong since it is tagged by machine rather than manually labeled.
Fortunately, Kay et.al \cite{kay2017kinetics} construct a large-scale and well-labeled benchmark, called Kinetics,
which covers 400 human action classes with at least 400 video instances per action category.
To activate as many neurons as possible in the 3D-CNN, we learn the parameter of the T-C3D on Kinetics with the temporal encoding method.
Experiments demonstrate that pre-training on large-scale datasets significantly improves accuracy.

\subsection{Aggregation Functions}
As shown in Equation \ref{equ:03} , aggregation functions are one of the key components in the T-C3D architecture.
In this subsection, we insightfully describe and analyze four fusion functions, e.g., mean pooling, maximum pooling, weighted pooling, and attention pooling.

\textbf{Average Pooling.} The basic hypothesis of mean pooling is to adopt the outputs of all clips equally for recognizing actions.
However, a few videos contain redundant or non-discriminative clips that are unrelated to the events,
Therefore, integrating these noisy clips might not precisely depict the event character.

\textbf{Maximum Pooling.} The intuition of maximum pooling is to select the most powerful clip for each action class
and represent the multiple clips with this strongest activation.
In some cases, one clip is not discriminative enough to distinguish these similar actions.
Thus, T-C3D degrades to the prior work which learns the network weights with one clip instance per video.

\textbf{Weighted Pooling.} This aggregation module aims to yield a serious of linear weights to operate
element-wise weighted linear integration over the outputs of each clip.
We employ this fusion method based on the observation that some action is formed by several stages and these stages may have different effects on recognizing actions.
In our experiment, we implement the function using a convolutional layer with the kernel of $S\times1$.

\textbf{Attention Pooling.} This aggregation module borrows the memory attention block from the memory network \cite{sukhbaatar2015end}.
The basic idea therein is to utilize a neural model to read external memories through a differentiable addressing/attention scheme.
In our experiment, we consider the outputs of each clip as the memory and cast feature weighting as a memory addressing procedure.

Formally, let ${F^{s}}$ be the 3D-CNN feature map of $s^{th}$ clip, then the fusion function filters them with a kernel $q$ via dot product,
generating a serious of corresponding weights ${e_{s}}$.
Then we conduct the softmax operator on them to produce normalized weights ${\omega_{s}}$ with $\sum\limits_{s=1}^{S}\omega_{s} = 1$.
The above two procedures can be formulated as the Equation \ref{equ:05}
\begin{equation}
   \omega_{s} =\frac {\exp (q^{T} {F_s})} {\sum\limits_{j=1}^{S}\exp (q^{T} {F_j})}.
\label{equ:05}
\end{equation}

\subsection{Deep Compression.} As shown in Figure \ref{fig:02}(b), we deploy the deep compression technique to reduce the model size after the T-C3D is fully trained.
The model trained with the above temporal encoding method is the input of deep compression.
This algorithm consists of the following five steps: sparsification, fine-tuning, pruning, quantization, Huffman Coding.
Then, we will describe each step in detail.

\textbf{Sparsification.} Inspired by \cite{han2016dsd}, we employ the dense-sparse-dense strategy to further increase the capacity of T-C3D.
In this work, we extend this idea to 3D-CNN for action recognition by performing the Sparse and Dense procedures.
In essence, the Sparse operation is pruning the weights of T-C3D.
More specially, in each layer (both the convolution layers and fully-connected layers), the weights are sorted by the absolute value.
Next, we set a single hyperparameter: the sparsity, the percentage of weights.
At the beginning of this procedure, the sparsity is 0.
We progressively prune the smallest weights to achieve the specified sparsity.
For example, if the sparsity is set to $0.8$, the sparsity will add $0.1$ after one epoch.
The whole optimization procedure is stopped at $8th$ epoch.
In this procedure, we set the sparsity to 0.9 for fully-connected layers and 0.85 for convolutional layers.

\textbf{Fine-tuning.} After obtaining the pruned model, we initialized these previously-pruned weights as zero and
the entire network is trained with a smaller learning rate (1/10 the original learning rate).
In other words, we progressively subtract the sparsity until it is 0.
To some degree, the function of the sparse-dense strategy is similar to the dropout.
The Sparse step might eliminate redundancy in 3D-CNN while the Dense procedure possibly arrives at a better local minimum.
The combination of sparsification and fine-tuning can increase the capacity of models and improve the recognition performance.
More details about experimental results are shown in Table \ref{tab:10}.

\textbf{Pruning.} After performing the Sparse and Dense operations, we prune the network weights to get the sparse model.
The operating steps are the same as the Sparse strategy.
In this procedure, we set the sparsity to $0.95$ for fully-connected layers and 0.9 for convolutional layers.
Therefore, the pruning procedure reduces the number of parameters by $10 \times$ for T-C3D model.

Following \cite{han2015deep}, we adopt the compressed sparse column (CSC) format to store the sparse model weights.
This format requires $2a+m+1$ numbers, where $a$ is the number of non-zero elements and $m$ is the number of columns.

To further compress the T-C3D, we save the index difference rather than the absolute position,
and encode this difference with 8 bits for the convolutional layer and 5 bits for the fully-connected layer.
When we need an index difference larger than the bound, we employ the zero-padding scheme.
For example, when we encode weights with 4-bit and the difference exceeds 16, then we add a filler zero.
More importantly, we can employ some library designed for optimizing the sparse matrix-vector multiplication (e.g., cuSPARSE) to further accelerate the T-C3D.

\textbf{Quantization.} We adopt the network quantization and weight sharing strategy to compress the pruned model.
In essence, this scheme adopts the network quantization to reduce the number of bits required to represent each weight and
weight sharing to reduce the weights required to represent.
We minimize the number of discriminative weights required to store by sharing the same weight among multiple connections and then fine-tune those shared weights.

First of all, the weights of T-C3D are clustered to several bins, then the weights in the same bin share the one value.
Therefore, for every weight, we only need to store a small index into a table of shared weights.
During the training phase, all the gradients with the same bin sum together,
then multiply by the learning rate and subtract from the shared centroids from the last iteration.

In general, for $k$ clusters (e.g., $k$ shared weights), we only need $log_2(k)$ bits to represent the index.
Given a model with $n$ connections and each connection is encoded with $b$ bits, constraining the connections to have only $k$ shared weights,
the compression rate $r$ can be formulated as:
\begin{equation}
   r =\frac {nb} {nlog_2(k) + kb}.
\label{equ:06}
\end{equation}

In our experiment, a $k$-means clustering algorithm is employed to group the weights for each layer of the fully trained T-C3D.
Then the weights that quantized into the same group will share the same weight.
Please note that the weights are not shared across layers.
For the pruned T-C3D, we can quantize each convolutional layer using 8-bits (256 shared weights),
and 5-bits (32 shared weights) for each fully-connected layer with a slight loss of accuracy.

\textbf{Huffman Coding.}
A Huffman code is an optimal prefix code widely adopted for lossless data compression.
It employs variable-length codewords to encode source symbols.
The table is derived from the occurrence probability for each symbol.
More common symbols are represented with fewer bits.
Thus, we can employ Huffman Coding to further compress the model size of T-C3D.

We present the detail of this process in Algorithm \ref{alg:Framwork}.
In this work, the values and indexes of model weights are the source symbols.
First, we arrange symbols in descending order of probabilities.
Then, we merge two symbols of the lowest probabilities into one subgroup.
Next, we assign zero and one to top and bottom branches, respectively.
We conduct the above step until there is not more than one unmerge node.
Finally, we read transition bits on the branches from top to bottom
to generate codewords.
Experimental results demonstrate that Huffman coding these non-uniformly distributed values
(weights and indexes) saves more than 20\% of model storage.

  \begin{algorithm}[htb]
  \caption{Compress the model with Huffman Coding.}
  \label{alg:Framwork}
  \begin{algorithmic}[1]
    \Require
      the values and indexes of model weights
    \State arrange symbols in descending order of probabilities
    \Repeat
    \State merge last two symbols into one subgroup
    \State assign zero and one to the top and bottom branches
    \Until there is not more than one unmerge node
    \State read transition bits on the branches to generate codewords
    \Ensure
    codewords for each symbol
  \end{algorithmic}
\end{algorithm}

\section{APPLICATIONS}
\label{APP}

\subsection{Application I: Anomaly Recognition in Surveillance Videos}
\label{sbusec:videoc}
Recognizing anomaly in surveillance video \cite{Liu2017Generalized,liu2019exploring} can stop illegal activities and guarantee public safety.
For example, if the algorithms can detect the anomaly, e.g., shooting or arson, many lives can be saved and the property will be protected.
As its major functionality, our T-C3D framework can support anomaly recognition for the public security department.

Recently, Sultani et al. \cite{sultani2018real} construct a novel anomaly detection dataset that consists of 1,300 videos recorded from real-world surveillance cameras.
To prove our method can achieve excellent accuracy, we test our method on this dataset through the standard evaluation criteria.
As shown in Table \ref{tab:08}, our method achieves competitive performance with the state-of-the-art approaches.

In addition to the impressive recognition performance, the proposed framework also has two evident advantages over the current methods: speed and storage size,
which are indispensable for popular applications.
In our work, the final version of T-C3D can process the videos at $969$ frames per second which is higher than real-time by a large margin.
Our method reduces the size of T-C3D by 25MB from 127MB without degradation of accuracy. This allows
fitting the model into on-chip SRAM cache rather than off-chip
DRAM memory.
Our compression method also facilitates the deployment of complex neural networks in mobile
applications where application size and download bandwidth
are constrained.

\subsection{Application II: Video Classification}
The proposed T-C3D framework can also be applied to tackle the problem of video classification.
With the development of the Internet, people tend to utilize the videos to share their life on the Internet.
However, most videos on the website are not tagged or wrong-labeled.
Consequently, the video-sharing website, such as YouTube, can make use of the proposed method to assign the label for each video.
Then the consumer can retrieve the videos by searching the word of the label.
To evaluate the video classification accuracy of T-C3D, we conduct extensive experiments on three action recognition datasets,
including trimmed video datasets and untrimmed video benchmarks.
In essence, the task of action recognition is the problem of video classification.
Experimental results on these three datasets demonstrate that our method can fulfill the task of video classification in practice.
\newline
\newline
\newline

\section{Experiments}
\label{exper}
In this section, we first present the benchmarks and implementation details of T-C3D.
Next, we conduct extensive experiments to explore suitable alternatives for training T-C3D, such as integration module, computation complexity, and processing speed. Finally, we provide the comparison between our method and the state-of-the-art methods on classification accuracy, speed, and model size.
\subsection{Datasets and Evaluation Protocol}
\textbf{Action recognition.} We empirically evaluate our T-C3D framework on two popular action recognition datasets: UCF101 \cite{soomro2012ucf101}, HMDB51 \cite{kuehne2011hmdb}.

The UCF101 dataset consists of $13,320$ videos covered $101$ action classes.
Most videos in UCF101 have the $320\times240$ spatial resolution with $25$ Frames Per Second (FPS) frame rate.
Every action category has at least $70$ training examples.
The HMDB51 is a large-scale collection of real-world videos from a variety of sources, e.g., movies and video-sharing websites.
This benchmark is constituted by $6,766$ video clips in total from $51$ action classes.
Each prescribed split in HMDB51 contains $3,570$ training examples and $1,530$ test instances, while an official split in UCF101 is composed of around $9,500$ training and $3,700$ testing instances.

For UCF101 and HMDB51, we utilize the official training/testing splits provided in original work and average the accuracy over these splits as the evaluation scheme. Given the GPU memory limits, we only conduct the exploration study on the first split of UCF101.
In terms of the judgement of speed, the widely used FPS is adopted as metric and experiments are performed on a Tesla K40 GPU and a CPU (E5-2620 v4).

\textbf{Anomaly detection.} Recently, Sultani et al. \cite{sultani2018real} construct a novel anomaly detection dataset that consists of $1,900$ untrimmed videos
recorded from real-world surveillance cameras, with $13$ realistic anomalies and normal actions.
Following \cite{sultani2018real}, we adopt the frame-based Receiver Operating Characteristic (ROC) curve and the corresponding area under the curve (AUC)
as the evaluation metric.

\textbf{Action similarity label.} The ASLAN dataset includes $3,631$ videos from $432$ action categories.
This dataset is constructed to solve the problem of whether a given pair of videos contain the same action.
We follow the prescribed 10-fold cross-validation with the data splits provided with this dataset.
\subsection{Implementation Details}
No matter which task, each video is first parsed to produce the clips as the input of the network.
Besides, we employ two kinds of data augmentation tricks to minimize the influence of severe over-fitting.
The first technique is randomly flip half of clips.
The other trick extends the random crop with scale jittering and aspect ratio jittering techniques that are widely adopted in image recognition.
More specially, we randomly select the width and height of the cropped region on three scales 1, 0.875, and 0.75, yielding more training instances.
Next, all the cropped regions are resized into $112\times112$. Thus, the network adopts an $8\times112\times112$ input,
the largest that can fit within GPU memory limits and maintain a large enough mini-batch.
The network weights are trained in an end-to-end manner with the mini-batch stochastic gradient descent algorithm, where the momentum is 0.9 and the batch size is 8.

On the action recognition, we first train the network on the Kinetics \cite{kay2017kinetics} to initialize network weights.
On UCF101 and HMDB51, we randomly initialize the last fully-connected layer and add a dropout module after the global average pooling layer
with a high dropout ratio (set to 0.8 in experiments) to avoid over-fitting.
The initial learning rate is set to $0.005$ and decreased to its 1/10 after $8,000$ iterations and $15,000$ iterations.
The whole optimization procedure is stopped at $20,000$ iterations.
For HMDB51, the training strategy is the same as that of UCF101, except that the iteration numbers are adjusted according to the number of training instances.

On the anomaly detection, we consider it as an untrimmed video classification task because we take the anomalous video as one category
and normal videos as the other category.
On the action similarity label, we only adopt the T-C3D as a feature extractor to obtain the video visual features.
Specially, we extract the outputs of three layers in T-C3D, i.e., res5b, pool5, and prob as the visual representation of clips.
Then we average all clip-level features and L2-normalize each feature.
Following the protocol of \cite{kliper2012action}, we adopt the strategy of 10-fold cross-validation with the provided data splits on this dataset.

\subsection{Exploration Study}
\label{exper:4}
In this subsection, we perform extensive exploration study of the proposed framework from the following four perspectives:
1) sampling approaches for producing clips, 2) the number of clips sampled from one video,
3) integration function module, and 4) weights initialization strategy.
Please note that all experiments in this subsection are conducted on the split 1 of UCF101 dataset.

\textbf{Study on sampling methods.} We study the influences of two sampling schemes presented in Section \ref{sbusec:VideoComponentsGeneration}.
Feeding network consecutive frames obtains higher accuracy than sampling non-consecutive frames (89.5\% vs 89.2\%).
Although each clip generated by the latter sampling scheme can cover the entire video,
the two adjacent frames might contain quite a large-displace movement.
The 3D-CNN possibly fail to model large motion information.

\textbf{Study on aggregation functions.} Table \ref{tab:02} presents the results of different aggregation functions. The effect of the pooling operation on the results are analysed as follows.

The mean pooling adopts the outputs of all
clips equally for recognizing actions and utilize their average
activations to produce the whole video score. From this view,
mean pooling not only depicts sequences of clips but also
generates the visual feature of the overall video.

Maximum pooling concentrates on only one clip without considering the outputs of other clips.
In some cases, one clip is not discriminative enough to distinguish these similar actions.
As a result, this aggregation method drives the T-C3D to depict the whole video just with
one clip, which violates the T-C3D's original idea of capturing long-term information.

We employ weighted pooling and attention pooling based on the observation that some action is
formed by several stages and these stages may have different effects on recognizing actions.
To some extent, these functions can be regarded as the combination of the merits of maximum
pooling and average pooling, concentrating on the representation of discriminative clips while reducing the negative influence of clips.

Surprisingly, both weighted pooling and attention pooling do not increase the performance substantially.
This is partly because these datasets are clean with less irrelative background clips.
Thus, the simple aggregation function also can lead to good classification performance.
Concerning this observation, we adopt average pooling as the default aggregation function in the following sections.

\renewcommand\arraystretch{1.2}
\begin{table}[t]
\begin{center}
\centering \caption{Exploration of different aggregation functions for T-C3D on split 1 of UCF101 dataset.}
\label{tab:02}
\vspace{1mm}
\begin{tabular}{|l|c|c|c|c|c|}
  \hline
  Aggregation Functions & Accuracy   \\
  \hline
  \rule{0pt}{10pt}
  Max pooling      &88.1    \\
  \rule{0pt}{10pt}
  Average pooling      &89.4    \\
  \rule{0pt}{10pt}
  Weighted pooling      &89.1    \\
  \rule{0pt}{10pt}
  Attention pooling     &89.5    \\
  \hline
\end{tabular}
\end{center}
\end{table}

\renewcommand\arraystretch{1.2}
\begin{table}[t]
\begin{center}
\centering \caption{Exploration of different number of sampled clips for T-C3D on split 1 of UCF101 dataset.}
\label{tab:13}
\vspace{1mm}
\begin{tabular}{|p{0.3\columnwidth}<{\centering}|p{0.2\columnwidth}<{\centering}|} 
  \hline
  Snippets Number & Accuracy   \\
  \hline
 \rule{0pt}{10pt}
  1      &85.7    \\
 \rule{0pt}{10pt}
  3      &89.4    \\
 \rule{0pt}{10pt}
  6     &89.6    \\
 \rule{0pt}{10pt}
  8     &89.7    \\
  \hline
\end{tabular}
\end{center}
\end{table}

\renewcommand\arraystretch{1.2}
\begin{table}[t]
\begin{center}
\centering \caption{Evaluation of different parameter initialization schemes for T-C3D on split 1 of UCF101 dataset.}
\label{tab:04}
\vspace{1mm}
\begin{tabular}{|l|c|c|c|c|c|}
  \hline
  Parameter Initialization & Accuracy   \\
  \hline
  \rule{0pt}{10pt}
  Training on scratch (share weights)         &68.3    \\
  \rule{0pt}{10pt}
  Pre-train on Sports-1M (share weights)      &89.5    \\
  \rule{0pt}{10pt}
  Pre-train on Kinetics (share weights)       &92.5    \\
  \rule{0pt}{10pt}
  Pre-train on Kinetics (not share weights)   &91.4    \\
  \hline
\end{tabular}
\end{center}
\end{table}

\renewcommand\arraystretch{1.2}
\begin{table}[t]
\begin{center}
\centering \caption{Comparison of speed and accuracy based on different clip numbers and multi-scale strategies for T-C3D on split 1 of UCF101 dataset.}
\label{tab:05}
\vspace{1mm}
\begin{tabular}{|c|c|c|c|c|c|}
  \hline
   Input           & Accuracy  & FPS   & FLOPS($10^{10}$)\\
  \hline
  \rule{0pt}{10pt}
  All clips per video (multi-scale)   &92.8     &45         &96.5\\ 
  \rule{0pt}{10pt}
  S clips per video (multi-scale)     &92.2     &197        &22.1\\
  \rule{0pt}{10pt}
  All clips per video                 &92.5     &220        &19.7\\
  \rule{0pt}{10pt}
  S clips per video                   &91.8     &969        &4.4\\
  \hline
\end{tabular}
\end{center}
\end{table}

\textbf{Study on snippets number.} We explore the effect of the number of clips sampled from each video.
As shown in Table \ref{tab:13}, recognition accuracy is increasing when raising the number of clips.
Please note that the T-C3D weights are optimized without the temporal encoding algorithm when the number of clips is 1.
According to Table \ref{tab:13}, we can easily observe that the temporal encoding method can learn the discriminative feature for action representation.
We also see that there is only a slight improvement in accuracy when the number of clips increases from 3 to 8.
Concerning the balance between training time and accuracy, we sample 3 clips from each video in the later experiments.

\textbf{Study on parameter initialization.} We list the results of different parameter initialization schemes in Table \ref{tab:04}.
Obviously, initializing the T-C3D weights on a large-scale dataset can increase the performance by a large margin.
Compared with Kinetics, the Sports-1M dataset is not well labeled but has more training samples and more categories.
Thus, we can conclude that the quality is more critical than quantity when initializing the weights on a large-scale dataset.
Besides, we share weights of networks during training as this strategy not only reduces the model size but also avoids overfitting.
According to Table \ref{tab:04}, sharing weights achieves better performance due to fewer parameters.

\textbf{Study on the balance between speed and accuracy.} At the testing stage, to find the trade-off between the speed and accuracy,
we also investigate both multi-scale testing strategy and a single-center crop predicting scheme.
As shown in Table \ref{tab:05}, feeding the network with 5 crops and their corresponding mirror (multi-scale) slightly increases classification accuracy
but seriously degrade the speed.
Besides, we also yield the final video score with two methods: integrating all clips of the video or just fusing $S$ clips per video.
Owing to the temporal encoding method, which trains the network with sparse sampling,
only sampling $S$ clips from one video thus demonstrates competitive performance at fast speed.
Moreover, we also show another meaningful evaluation of floating point operations (FLOPs) in Table \ref{tab:05}.
Following \cite{tran2017convnet}, we only calculate FLOPs of networks forward process because the complexity of preprocessing of videos (extract videos into frames) is rather low.
Given the same input size, we observe that the setting obtained the high FPS always achieves the low FLOPs.

\subsection{Comparison with The-State-of-The-Art Methods}
In this section, we compare the performance of T-C3D with the state-of-the-art methods on four datasets; UCF101 \cite{soomro2012ucf101},
HMDB51 \cite{kuehne2011hmdb}, UCFCrime \cite{sultani2018real}, ASLAN \cite{kliper2012action}.
\renewcommand\arraystretch{1.35}
\begin{table*}[htbp]
\linespread{1.5}
\begin{center}
\newcommand{\tabincell}[2]{\begin{tabular}{@{}#1@{}}#2\end{tabular}}
\centering \caption{Comparison of accuracy, speed and model size with the state-of-the-art methods on UCF101 and HMDB51.}
\label{tab06}
\begin{tabular}{|p{0.2\columnwidth}<{\centering}|p{0.53\columnwidth}<{\centering}|p{0.32\columnwidth}<{\centering}|p{0.117\columnwidth}<{\centering}|p{0.137\columnwidth}<{\centering}|p{0.1\columnwidth}<{\centering}|p{0.16\columnwidth}<{\centering}|}
  \hline
  & Method    &Pre-train Dataset    &UCF101 &HMDB51    &FPS  &Model Size(/M)\\
  \hline
  \multirow{3}{*}{\tabincell{c}{Hand-crafted \\ Feature}}
  &improved Dense Trajectory+FV \cite{wang2013action} &None      &85.9   &57.2      &2 &N/A\\
  &Dense Trajectory+MVSV \cite{cai2014multi}       &None       &83.5   &55.9      &N/A    &N/A\\
  &Motion Vector+FV \cite{kantorov2014efficient}    &None  &78.5   &N/A       &133   &N/A\\
  \hline
  \multirow{6}{*}{\tabincell{c}{One-stream \\ (RGB)}}
  &C3D \cite{tran2015learning}          &Sports-1M          &82.3   &51.6      &314    &300\\
  &C3D(3nets) \cite{tran2015learning}   &Sports-1M          &85.2   &N/A       &\textless314    &900\\
  &Slow Fusion \cite{karpathy2014large} &ImageNet   &65.8   &N/A       &N/A    &233\\
  &Res3D \cite{tran2017convnet}         &Kinetics       &85.8   &54.9      &220    &127\\
  &Temporal Segment Network(RGB) \cite{wang2016temporal}    &ImageNet          &85.1   &51.0      &N/A    &39\\
  &P3D \cite{qiu2017learning}          &Sports-1M          &88.6   &N/A       &141    &98\\
  &3D-ResNet18 \cite{hara2017can}      &Kinetics   &84.4    &56.4 &220 &252 \\
  &3D-ResNet101 \cite{hara2017can}      &Kinetics   &88.9   &61.7 &\textless220 &653 \\
  &3D-ResNet152 \cite{hara2017can}      &Kinetics  &89.6   &62.4 &\textless220 &898 \\
  \hline
  \multirow{9}{*}{\tabincell{c}{Two-stream \\ (Based)}}
  &Two-stream(VGG-M) \cite{simonyan2014two}    &ImageNet  &88.0   &59.4      &14    &658\\
  &Two-stream(ResNet50) \cite{simonyan2014two} &ImageNet        &91.7   &61.2       &\textless14    &184\\
  &Two-stream+LSTM \cite{yue2015beyond}        &ImageNet     &88.6   &N/A       &\textless14    &105\\
  &(Two-stream+C3D)+LSTM  \cite{li2016action}
  &ImageNet+Sports-1M              &90.8   &63.6       &\textless14    &958\\
  &LTC \cite{varol2017long}             &Sports-1M            &91.7   &64.8      &\textless14    &305\\
  &AdaScan \cite{kar2016adascan}        &ImageNet             &89.4   &54.9      &\textless14    &1026\\
  &ActionVLAD \cite{girdhar2017actionvlad} &ImageNet                 &92.7   &66.9      &\textless14    &1026\\
  &TDD+FV \cite{wang2015action}          &ImageNet   &90.3   &63.2      &\textless14    &691\\
  &Temporal Segment Network(TSN) \cite{wang2016temporal} &ImageNet            &\textbf{94.2}   &\textbf{69.4}      &5    &79\\
  &Enhanced Motion Vector  \cite{zhang2016real}        &ImageNet        &86.4   &N/A       &390    &1037\\

  \hline
  \multirow{3}{*}{T-C3D}
  &\textbf{Ours(Kinetics)}            &Kinetics          &\textbf{92.5}   &\textbf{62.4}       &\textbf{220}    &\textbf{127}\\
  &\textbf{Ours(Fast)}                &Kinetics      &\textbf{91.8}   &\textbf{62.8}       &\textbf{969}    &\textbf{127}\\
  &\textbf{Ours(Final)}                &Kinetics    &\textbf{91.8}   &\textbf{62.8}       &\textbf{969}    &\textbf{4}\\
  \hline
\end{tabular}
\end{center}
\end{table*}

\textbf{Results on the UCF101 and HMDB51 datasets.}
The comparison of accuracy, speed, and model size is summarized in Table \ref{tab06}.
These methods can be briefly grouped into three categories:

\begin{enumerate}
    \item Hand-crafted feature methods: this kind of method always consists of three steps, e.g., designing feature,
    feature encoding, classification. Most of them often employ the SVM as the classifier.
    Thus, we use the format ``designed feature with feature encoding algorithm'' to represent this kind of approach:
    improved Dense Trajectory with Fisher Vector \cite{wang2013action}, Dense Trajectory with Multi-view super vector (MVSV) \cite{cai2014multi},
    and Motion Vector (MV) with Fisher Vector \cite{kantorov2014efficient};
    \item One-stream methods: this kind of method only adopts RGB frames as input, some methods utilize 3D convolutional filter with the input of several RGB frames,
    e.g., C3D \cite{tran2015learning}, Res3D \cite{tran2017convnet}, P3D \cite{qiu2017learning}, 3D-ResNet152 \cite{hara2017can}.
    Meanwhile, some approach adjust the architecture of neural networks, such as Slow Fusion \cite{karpathy2014large},
    and Temporal Segment Network (TSN) with input RGB \cite{wang2016temporal};
    \item Two-stream methods: The original two-stream method \cite{simonyan2014two} are fed with both single RGB frame and stacked optical flows.
    A lots of improved version are proposed: Two-stream+LSTM \cite{yue2015beyond}, LTC \cite{varol2017long}, AdaScan \cite{kar2016adascan},
    ActionVLAD \cite{girdhar2017actionvlad}, TDD+FV \cite{wang2015action} and TSN \cite{wang2016temporal}, and Enhanced MV \cite{zhang2016real}.
\end{enumerate}

Among these three kinds of methods, the two-stream based methods achieve the highest classification accuracy with the slowest speed.
This is primarily due to the extraction of optical flow which is computationally expensive and contains fine structures.
Besides, whether one-stream or two-stream based method, it exceeds hand-crafted methods on the accuracy.
Most deep neural networks integrate the feature extraction and classification into one network, which is more convenient to deploy.
All of the two-stream based approaches are too computationally expensive to process videos at real-time because of the heavy calculation of optical flows.
Thus, both designing a light-weight medium to replace optical flow and improving the 3D convolution network are promising directions.

Compared with hand-crafted features, T-C3D exceeds the most powerful manually designed feature (iDT) encoded with Fisher Vector.
Besides, it achieves the highest accuracy and fastest speed among one-stream based methods which only utilize RGB frames as input.
Especially, T-C3D consists of 18 layers, but it can achieve superior accuracy with the deep 3D-CNNs,
such as 3D-ResNet101 \cite{hara2017can} and 3D-ResNet152 \cite{hara2017can}.
Pre-training strategy indeed is significant for the action representation and
we report the pre-training dataset for each method in Table \ref{tab06}.
According to Table \ref{tab06}, Res3D \cite{tran2017convnet}, 3D-ResNet101\cite{hara2017can}
and 3D-ResNet152 \cite{hara2017can} adopt the same pre-training dataset as ours.
However, our approach can achieve superior accuracy on both benchmarks with the faster speed and smaller model size.
According to the latest results of TSN, it presents competitive accuracy of $87.3\%$ in 340 fps speed on UCF101 split 1 when using RGB and RGB Difference.
However, compared with our method, its accuracy is lower and its computational complexity is higher.

Compared with two-stream based methods, our method achieves better recognition performance than the original two-stream architecture even with an extra deep CNN network.
More importantly, it attains competitive accuracy that is superior or very close to some recently published work, such as AdaScan \cite{kar2016adascan} and ActionVLAD \cite{girdhar2017actionvlad}.
Despite outstanding results on HMDB51 and UCF101, the computation overhead of TSN is so expensive (5 fps) that it is difficult or impossible to apply in practice,
especially with the real-time demand.

For processing speed, our method significantly outperforms the state-of-the-art real-time method.
Especially, the final version of T-C3D can achieve the $969$ FPS.
As for model size, after performing the deep compression,
our method beats other approaches by a large margin without sacrificing the classification accuracy,
which dramatically promotes the application of action recognition in the real-world.

\renewcommand\arraystretch{1.6}
\begin{table}[t]
\begin{center}
\centering \caption{Comparison of AUC with the state-of-the-art methods on UCFCrime dataset for anomaly detection task.}
\label{tab:08}
\vspace{1mm}
\begin{tabular}{|p{0.45\columnwidth}<{\centering}|p{0.14\columnwidth}<{\centering}|}
  \hline
  Method  & AUC \\   \hline
  C3D(Binary classifier) \cite{tran2015learning}            &50.0\\ \hline
  Hasan et al. \cite{hasan2016learning}     &50.6\\ \hline
  Lu et al. \cite{lu2013abnormal}         &65.5\\ \hline
  Multiple Instance Learning \cite{sultani2018real}       &75.4\\ \hline
  \textbf{Ours}                          &\textbf{76.3}\\ \hline
\end{tabular}
\end{center}
\end{table}
\textbf{Results on the UCFCrime dataset.} On this dataset, we evaluate our method on the task of anomaly detection.
This task is quite challenging because the videos are long and untrimmed.
In other words, we only know the training videos are normal or not, but we do not know when the anomaly starts and ends.
In our experiments, we regard the anomaly detection as a binary classification while the anomalous activity recognition as a standard multi-class classification.

In Table \ref{tab:08}, we can observe that our T-C3D outperforms all state-of-the-art methods for anomaly detection.
More specially, T-C3D improves the very recent method in \cite{sultani2018real} from 75.4\% to 76.3\%,
which adopts the multiple instance learning algorithm \cite{wang2018revisiting,tang2018pcl} to detect anomaly in surveillance videos.
Unlike the performance on UCF101 and HMDB51 for video-based action recognition task,
C3D fails to detect the anomaly since it achieves the same accuracy as the random choice.
However, T-C3D can successfully detect the anomaly owing to its capacity to capture the long-term information.
This suggests that T-C3D cannot only solve the real-world problem well but also achieve remarkable performance on the challenging untrimmed video datasets.

Compared with the performance of previous work on UCF101 and HMDB51, popular datasets for recognizing daily normal actions,
anomalous activity recognition is more challenging due to its unique complexities, e.g., low resolution, dark night condition.
Thus, designing methods for anomalous activity recognition should consider these unique properties.

\renewcommand\arraystretch{1.8}
\begin{table}[t]
\begin{center}
\centering \caption{Comparison of AUC with the state-of-the-art methods on ASLAN dataset.}
\label{tab:10}
\vspace{1mm}
\begin{tabular}{|p{0.45\columnwidth}<{\centering}|p{0.14\columnwidth}<{\centering}|p{0.14\columnwidth}<{\centering}|}
  \hline
  Method  & Model & AUC \\   \hline
  IDT+FV \cite{wang2013action}                       &metric &75.4\\ \hline
  ResNet152 \cite{he2016deep}                       &linear &77.4\\ \hline
  C3D \cite{tran2015learning}                        &linear &86.5\\ \hline
  P3D \cite{qiu2017learning}                        &linear &\textbf{87.8}\\ \hline
  \textbf{Ours}                                      &linear &\textbf{87.4}\\ \hline
\end{tabular}
\end{center}
\end{table}

\renewcommand\arraystretch{1.9}
\begin{table}[t]
\begin{center}
\centering \caption{Accuracy and model size of T-C3D with deep compression techniques.}
\label{tab:11}
\vspace{1mm}
\begin{tabular}{|p{0.25\columnwidth}<{\centering}|p{0.14\columnwidth}<{\centering}|p{0.18\columnwidth}<{\centering}|}
  \hline
  Method & Accuracy  & Model size  \\   \hline
  Baseline T-C3D       &92.5   &127 \\ \hline
  Sparse               &92.6   &127 \\ \hline
  Dense                &92.9   &127 \\ \hline
  Prune                &92.9   &13.5 \\ \hline
  Quantization         &92.4   &6.7 \\ \hline
  Huffman Coding       &92.4   &4.4 \\ \hline
\end{tabular}
\end{center}
\end{table}

\textbf{Results on the ASLAN dataset.} Table \ref{tab:10} summarizes the performance and comparison with the state-of-the-art methods on the ASLAN dataset.
Overall, our T-C3D performs consistently better than both hand-crafted feature representations and deep neural networks based approaches
using the official protocol of the area under the ROC curve (AUC).
In general, deep neural networks based methods exhibit better performance than hand-crafted feature representations.
Surprisingly, C3D outperforms the powerful 2D-CNN (ResNet-152) by a large margin on this challenging task.
This is partly because the action similarity labeling task requires more the scene or action features rather than the knowledge for recognizing some specific objects.
Compared with both shallow features and deep neural networks such as C3D \cite{tran2015learning}, we can observe the apparent improvements.
Besides, T-C3D contains 18 layers, but it achieves comparable performance with the P3D \cite{qiu2017learning} which consists of more than 200 layers.
This demonstrates that T-C3D is endowed with the advantages of both extra deep 2D-CNN and C3D.
\begin{figure*}[t]
\begin{minipage}[b]{0.96\linewidth}
  \centering
  \centerline{\includegraphics[width=18cm,height=9.2cm]{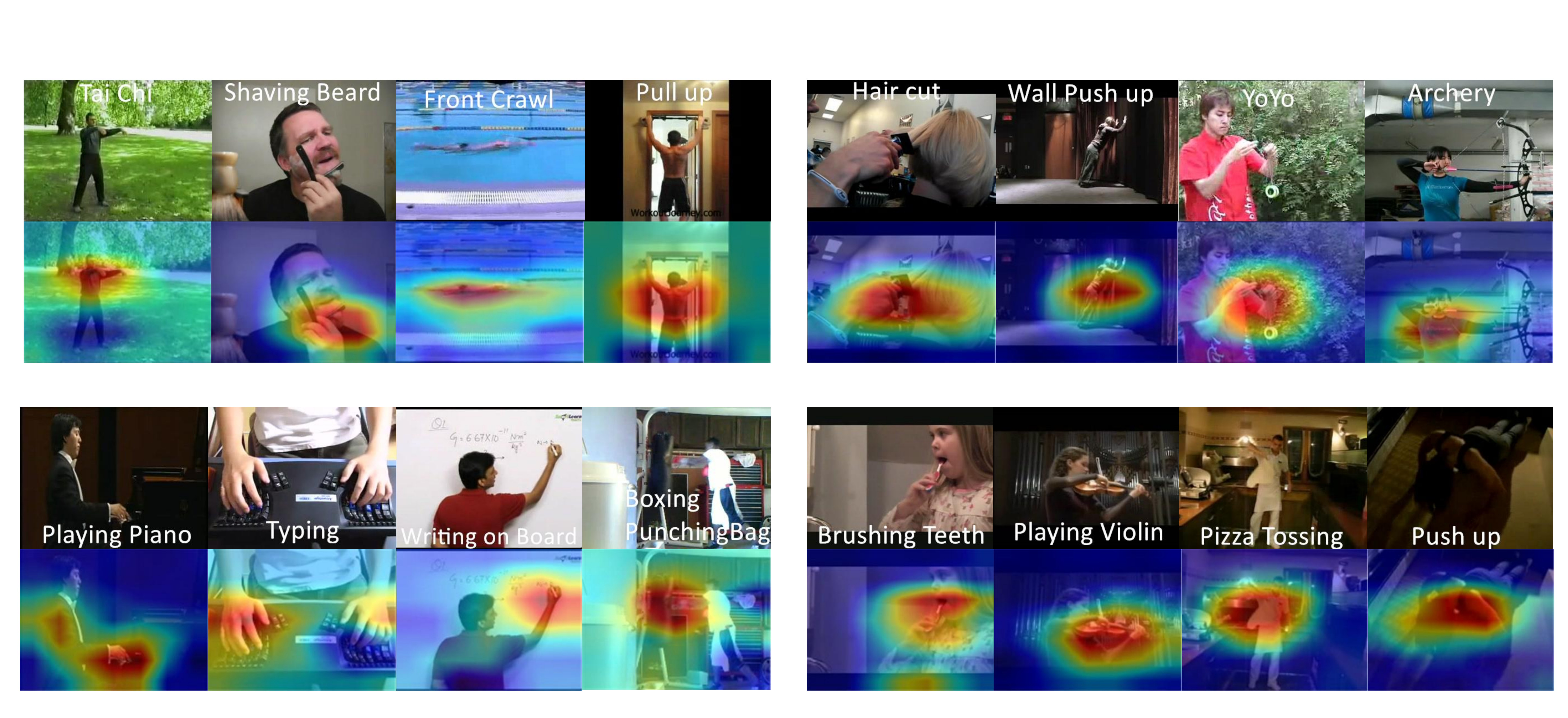}}
\end{minipage}
\centering \caption{Visualization of T-C3D models for action recognition using Class Activation Mapping (CAM) in \cite{zhou2016learning}. Best seen in color and with zoom.}
\label{fig:04}
\end{figure*}

\textbf{Effectiveness on deep compression.} After training the T-C3D, we make efforts to reduce the model size without influencing the accuracy.
Table \ref{tab:11} lists the classification accuracy and model size when performing the following five steps: sparse, dense, prune, quantization, Huffman Coding.
In all experiments about deep compression, we add a dropout layer after the global average pooling layer with a high dropout ratio (set to 0.8 in experiments).
Interestingly, we can observe that the sparse step increase accuracy due to its capacity to avoid over-fitting.
This suggests that the 3D-CNNs are always redundant and tends to over-fitting, leading to degradation of classification accuracy.
Quantization procedure decreases the performance because it is difficult to employ a small bit (8 bit) to represent the extensive information.
In summary, the model size is reduced progressively by performing specific procedures step by step.
According to the final version result, we significantly reduce the model size while preserving accuracy.

\subsection{Visualization Results.} Besides performance, we would like to provide in-depth analyses for T-C3D models.
In this sense, we adopt the technique of Class Activation Mapping (CAM) in \cite{zhou2016learning} to understand what T-C3D has learned for action recognition.
We can utilize this technique to visualize the predicted class scores on the given clip, highlighting the discriminative motion parts learned by the T-C3D.

We randomly pick sixteen categories from the UCF101 dataset, such as shaving beard, yoyo, playing guitar, hair cut, boxing punching bag and so on.
We present the RGB frame, RGB frame covered with heat-map results in Figure \ref{fig:04}.
The predicted categories are marked with white font on the RGB frame.
As illustrated in Figure \ref{fig:04}, T-C3D model could precisely localize the motion parts in both the small motion and long-term motion.
Take the category of playing the piano as an example, our model captures not only the interactions between the hands and piano but also the shook back.
The motion between the hands and piano is small while the back motion requires long-term information to be observed.
Besides, considering ``shaving beard'' and ``writing on board'' as the example,
it is also easy to notice that T-C3D predicts the right category through detecting discriminative human-object interactions accurately.
These above observations suggest that T-C3D models perform well on learning action representation,
which is also well reflected in our above quantitative experiments.

\section{Conclusion}
In this work, we first present T-C3D, an end-to-end trainable framework equipped with the temporal encoding method to model the multiple granularity features of videos.
Then, we integrate the deep compression techniques with the temporal encoding algorithm to further accelerate the employment of this framework.
Besides, T-C3D can process video at a faster speed than real-time.
Experiments conducted on four datasets over three tasks demonstrate the effectiveness of the discriminative action representation produced by T-C3D.
Performance improvements are clearly observed when comparing to the state-of-the-art real-time methods.
In the future, we will extend T-C3D for online processing to recognize the actions once the frames are received instead of presenting the entire video.
\newline

\section{Acknowledgement}
This work is partially supported by the Funds for International Cooperation and Exchange of the National Natural Science Foundation of China (No. 61720106007),
and the Funds for Creative Research Groups of China (No.61921003).
This work is also supported by the 111 Project (B18008) and BUPT Excellent Ph.D. Students Foundation (CX2018313).

\ifCLASSOPTIONcaptionsoff
  \newpage
\fi

\bibliographystyle{IEEEtran}
{
\bibliography{tcsvt}
}

\end{document}